\documentclass[utf8]{FrontiersinHarvard} 
\usepackage{url,hyperref,lineno,microtype,subcaption}
\usepackage{bm}
\usepackage{amsmath}
\usepackage{booktabs}
\usepackage[onehalfspacing]{setspace}
\usepackage{graphicx}
\usepackage{xcolor} 
\usepackage{changes} 
\usepackage{soul} 
\def\firstAuthorLast{Xiyun Li {et~al.}} 
\def\Authors{Xiyun Li$^{1,2}$, Ziyi Ni$^{1,3}$, Jingqing Ruan$^{1,2}$, Linghui Meng$^{1,3}$, Jing Shi$^{1,3}$, Tielin Zhang$^{1,3,*}$ and Bo Xu$^{1,2,3,4,*}$}
\def\Address{$^{1}$Institute of Automation, Chinese Academy of Sciences, Beijing, China \\
$^{2}$School of Future Technology, University of Chinese Academy of Sciences, Beijing, China \\
$^{3}$School of Artificial Intelligence, University of Chinese Academy of Sciences, Beijing, China \\
$^{4}$Center for Excellence in Brain Science and Intelligence Technology, Chinese Academy of Sciences, Shanghai, China }
\def\corrAuthor{Tielin Zhang and Bo Xu}
\def\corrEmail{tielin.zhang@ia.ac.cn, xubo@ia.ac.cn}
\begin{document}
\onecolumn
\firstpage{1}
\title {Mixture of personality improved Spiking actor network for efficient multi-agent cooperation}

\author[\firstAuthorLast ]{\Authors} 
\address{} 
\correspondance{} 
\extraAuth{}
\maketitle
\begin{abstract}
Adaptive human-agent and agent-agent cooperation are becoming more and more critical in the research area of multi-agent reinforcement learning (MARL), where remarked progress has been made with the help of deep neural networks. 
However, many established algorithms can only perform well during the learning paradigm but exhibit poor generalization during cooperation with other unseen partners. 
The personality theory in cognitive psychology describes that humans can well handle the above cooperation challenge by predicting others' personalities first and then their complex actions. Inspired by this two-step psychology theory, we propose a biologically plausible mixture of personality (MoP) improved spiking actor network (SAN), whereby a determinantal point process is used to simulate the complex formation and integration of different types of personality in MoP, and dynamic and spiking neurons are incorporated into the SAN for the efficient reinforcement learning. The benchmark Overcooked task, containing a strong requirement for cooperative cooking, is selected to test the proposed MoP-SAN. The experimental results show that the MoP-SAN can achieve both high performances during not only the learning paradigm but also the generalization test (i.e., cooperation with other unseen agents) paradigm where most counterpart deep actor networks failed. 
Necessary ablation experiments and visualization analyses were conducted to explain why MoP and SAN are effective in multi-agent reinforcement learning scenarios while DNN performs poorly in the generalization test.

\section{Keywords:} Multi-Agent Cooperation, Personality Theory, Spiking Actor Networks, Multi-Agent Reinforcement Learning, Theory of Mind.

\end{abstract}
\section{Introduction}
With the rapid development and great progress of deep reinforcement learning (RL) in recent years \citep{vinyals2019grandmaster,silver2017mastering,vaswani2017attention,yu2021surprising,meng2023offline}, more and more researchers have shown an increased interest in multi-agent cooperation, or human-in-the-loop cooperation \citep{shih2021critical,strouse2021collaborating,carroll2019utility,zhao2021maximum, shih2022conditional,ruan2022gcs,lou2023pecan}. 
However, cooperation with unseen partners usually requires continuous collection of expert data, which is expensive and delayed 
\citep{carroll2019utility,shih2022conditional}. 
Other methods attempt to achieve better generalization without expert data by constructing a population pool for simulating diverse candidate partners.
However, these works try to improve the generalization cooperation score by relying on being trained with a large number of well-designed partners but ignore the cultivation of the agent's real thinking and empathy ability. 

The less consideration of the psychological characteristics of partner agents might be the key reason why these artificial agents fail, compared to their counterpart biological agents. In our daily life, humans can cooperate well with others whom we have never seen before \citep{boyd2009culture,rand2013human}. This phenomenon is interesting but not hard to guess. We can infer others' personalities quickly, and then we can well handle the following cooperation behaviors with the help of this guessed personality. The personality theory is under the framework of theory of mind (ToM) \citep{gallagher2003functional,frith2005theory,aru2023mind,roth2022cutting}, which refers to our ability to speculate on the intentions, behaviors, and goals of other people, which explains why humans can collaborate with unseen partners from a cognitive perspective. In fact, instead of being classified into a specific personality, the unseen human can be viewed as some combination of several ``personalities." Therefore, it is significantly helpful to find as few representative personalities as possible and make them orthogonal to each other 
for a more efficient combination. The personality theory \citep{ryckman2012theories,schultz2016theories,mccrae2008five} from cognitive psychology has provided an opportunity to model the partners more clearly and concretely, including the big five personalities \citep{de2000big} and the sixteen personality factors (16PF) \citep{cattell2008sixteen}. These theories are useful in describing unique and diverse people\citep{anglim2021effect} and can instruct many cognitive tasks, such as personality trait tests \citep{o2007big}, to analyze people's suitable careers. 

Unlike the personality theory in cognitive science, which is often used as the discrete classification, we propose the base personality similar to the base vector in the personality space, which can be used for inferring personality. To further ensure the difference between multiple base personalities, determinantal point process (DPP) constraints are adopted as an intrinsic reward. Based on the personality model with these base personalities, the agent can naturally predict and understand any unseen partner to better make responses and obtain cooperation.

Hence, inspired by the above personality theory, we propose the mixture of personality (MoP), along with our previously proposed spiking agent network (SAN), which has been verified efficient in single-agent reinforcement learning \citep{zhang2022multi}. The SAN is biologically reasonable, containing more dynamic neurons, which have shown advantages in dynamic RL tasks with lower energy consumption and better generalization. Here we further applied SAN to MARL cooperation scenarios. Our main contributions can be concluded in the following parts:

\begin{enumerate}
\item We are the first to propose the concept of the MoP, which is inspired by the personality theory in psychology, describing a two-step prediction, where the personality estimator (PE) is designed to receive context for estimating the personality of partner under the DPP constraints first, and then behavior prediction is given by the multi-personality network.
\item We incorporate efficient SAN and MoP models to reach multi-scale biological plausibility, where spiking neurons with neural dynamics have been verified efficient in RL-like tasks \citep{zhang2022multi}, and we run further to combine neuronal scale dynamics and partner scale cooperations together, to increase the generalization ability of the agent in multi-agent collaboration.
\item The proposed MoP-SAN is then tested in the Overcooked benchmark environment, and the experimental results show a marked better generalization, especially when cooperating with other unseen partners compared to other DNN baselines, which means our proposed algorithm can successfully infer the personality of the unseen partner in the zero-shot collaboration test. We conducted analysis experiments to analyze why SAN method has better generalization results than DNN baselines. 
\end{enumerate}

\section{RELATED WORKS}
RL is an essential paradigm in machine learning, which is also suitable for many sequential decision-making tasks. The RL methods have recently achieved good results in many tasks \citep{vinyals2019grandmaster,silver2017mastering,silver2018general}. Existing traditional RL methods can be divided into value-based methods \citep{mnih2013playing} and policy-based methods \citep{schulman2015trust}. The proposal of actor-critic method is of milestone significance in RL which combines the advantages of value-based and policy-based methods. Proximal policy optimization (PPO) \citep{schulman2017proximal} is 
one of the most classic methods in this framework, which has achieved compelling performance in many tasks, such as control tasks \citep{schulman2017proximal} and StarCraft \citep{yu2021surprising}.

MARL describes the process of multi-agent learning strategies from scratch to maximize the global rewards in the process of interacting with the environment sequentially or simultaneously. For example, in the two-player cooperative task Overcooked, the ego agent and the partner agent need to cooperate to maximize the team reward from the Overcooked environment. In MARL, cooperative MARL tasks are a very challenging direction. Although there are some works exploring how to solve challenging problems in cooperative MARL tasks such as credit assignment\citep{rashid2020monotonic,sunehag2017value}, how to design a model which can generalize to unseen partners is still challenging.
For multi-agent cooperation, some recent works \citep{shih2021critical,carroll2019utility,zhao2021maximum,strouse2021collaborating,shih2022conditional,lou2023pecan} focus on the generalization research of unseen partners. Although traditional self-play methods \citep{silver2018general} have achieved significant advantages and can often converge to an optimal equilibrium strategy in competitive games, they tend to overfit specific partners for cooperative tasks. Some efforts are put into solving the overfitting through imitation learning \citep{carroll2019utility,shih2022conditional}, even though it has been reported as challenging in collecting expert data in many real scenarios. For the better generalization of human-AI collaboration, modular methods are proposed, which explicitly separate the convention-dependent representations and rule-dependent representations \citep{shih2021critical}. Other works \citep{strouse2021collaborating,zhao2021maximum} try to slove the cooperative task of unseen partners by designing various population pools, which include many carefully designed criteria and agents.

Since brain-inspired SNN has advantages in many aspects, many works have begun to use SNN to solve reinforcement learning problems \citep{florian2007reinforcement,tang2020reinforcement,zhang2022multi,bellec2020solution,patel2019improved,fremaux2013reinforcement}. Our previous work proposed a multi-scale dynamic coding improved spiking actor network (MDC-SAN) in a single-agent scenario to achieve efficient decision-making\citep{zhang2022multi}. Unlike most of these works that explore SNN methods in single-agent RL tasks, this paper wants to apply the SNN method to multi-agent cooperation tasks. In this work, we need to cooperate with different styles of partners in cooperative tasks, so it is vital to construct a model for partner modeling.

ToM \citep{gallagher2003functional,frith2005theory,aru2023mind,roth2022cutting} is a fundamental concept in cognitive psychology, and it allows individuals to predict and explain others' behaviors, communicate effectively, and better engage in cooperative interactions, which is also what we want AI agents to achieve. There are some works that design ToM models \citep{yuan2022situ,wang2021tom2c,tabrez2020survey} to solve RL tasks. Through the ToM model, the agent can communicate with other partners more efficiently and learn some conventions for partners. In some works \citep{rabinowitz2018machine,roth2022cutting}, the design of the ToM model is to understand the behavior of other agents, which is vital for many RL tasks. While ToM encompasses many aspects, including mental simulation, action prediction, and reasoning, in this context we will focus on a specific aspect called personality traits in order to enhance the agent model.

\section{Method}

\subsection{The problem setting of 2-player cooperation}

We can define this 2-player Markov game as a tuple $\left(\mathcal{O}, \mathcal{A}, \mathbb{P}, \mathbf{\gamma}, \mathbf{\pi}, \rho^{i}, r, m\right)$, where $\mathcal{O}$ denotes the observation space and $\mathcal{A}$ represents the action space that the ego agent and partner share. We can define $\mathbf{o}=(o^{1},o^{2})$ including the ego observation and the partner observation. We can denote label $\mathbf{a}=(a^{1},a^{2})$ as the joint actions for all players, including the ego action and the partner action. 
$\mathbb{P}: \mathcal{O} \times \mathcal{A} \rightarrow \mathcal{O}$ represents the environment transition probability function, and $\gamma$ $\in[0,1)$ is the discount factor. $\pi$ is the joint policy, and the policy of ego agent $\rho^{1}$ is the spiking policy of the SAN agent for our MoP-SAN, and $\rho^{2}$ represents the partner's policy. All agents share the same team reward function $\mathbf{r(o,a)}: \mathbf{o} \times \mathbf{a} \rightarrow R$. $\tau  = ({\mathbf{o}_0},{\mathbf{a}_0},{\mathbf{o}_1},...)$ denotes the trajectory generated by the joint policy $\pi$, and $\tau^{2}  = ({{o}_0^{2}},{{a}_0^{2}},{{o}_1^{2}},...)$ is the trajectory of the partner. The MoP model $m$ can model the partner based on the historical trajectory information of the partner and provide actionable guidance for the SAN agent. At each time step, the SAN agent perceives an observation $o_{t}^{1} \in \mathcal{O}$ and receives the guided action $\hat{a}_{t}^{2}$ from the MoP model $m$, taking action $a_{t}^{1} \in \mathcal{A}$ drawn from a spiking policy $\rho^{1}: \mathcal{O} \times \mathcal{A} \rightarrow[0,1]$, denoted as $a^1_t=\rho^1(\cdot | o_t^{1}, \hat{a}_{t}^{2})$. The policy of the partner can be denoted as $a^2_t=\rho^2(\cdot | o_t^{2})$. The SAN agent and partner enter the next state ${\mathbf{o}_{t+1}}$ with the probability $\mathbb{P}\left({\mathbf{o}_{t+1}} \mid {\mathbf{o}_t}, \mathbf{a_{t}}\right)$, receiving a numerical reward $r_{t+1}$ from the environment. All agents coordinate together for the maximum cumulative discounted return
${\mathbb{E}_{\bm{\tau} \sim \pi}}\left[ {\sum_{t = 0}^{\infty} {{\gamma ^t}r({{\mathbf{o}_t}},{\mathbf{a}_t})} } \right]$.

We assume that there is at least one joint policy through which all agents can attain the maximum cumulative rewards in fully cooperative games. The problem, objective statement, and our approach are formalized in the following sections.

\subsection{The algorithmic architecture and pipeline of MoP-SAN}

\begin{figure*}[htb]
  \centering
\includegraphics[width=16cm]{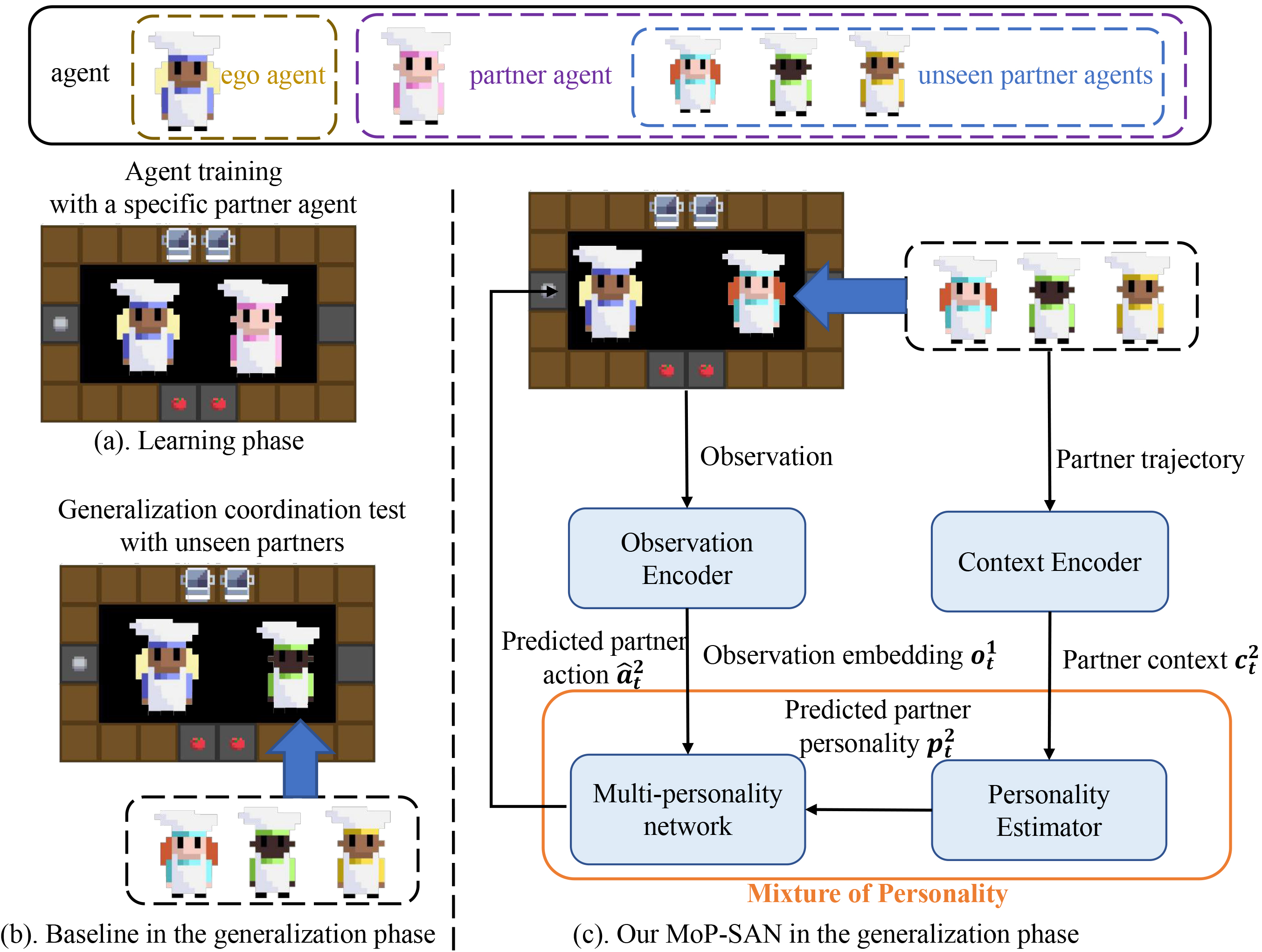}
  \caption{The learning and generalization phases of our proposed MoP-SAN. (a). The ego agent and the specific partner agent collaborate to complete the Overcooked task in the learning phase. (b). In the generalization phase, the ego agent needs to collaborate with some unseen partner agents to test generalization ability (zero-shot collaboration). (c). Our MoP-SAN also needs to collaborate with unseen partners. By constructing a MoP model, we can estimate partner personality first and predict the actions of the partner according to the personality of the partner. Two agents in the same kitchen in all three graphs represent the cooperative relationship between the two agents to complete this cooking task.}
  \label{fig:overview}
\end{figure*}

\begin{figure*}[htb]
  \centering
\includegraphics[width=16cm]{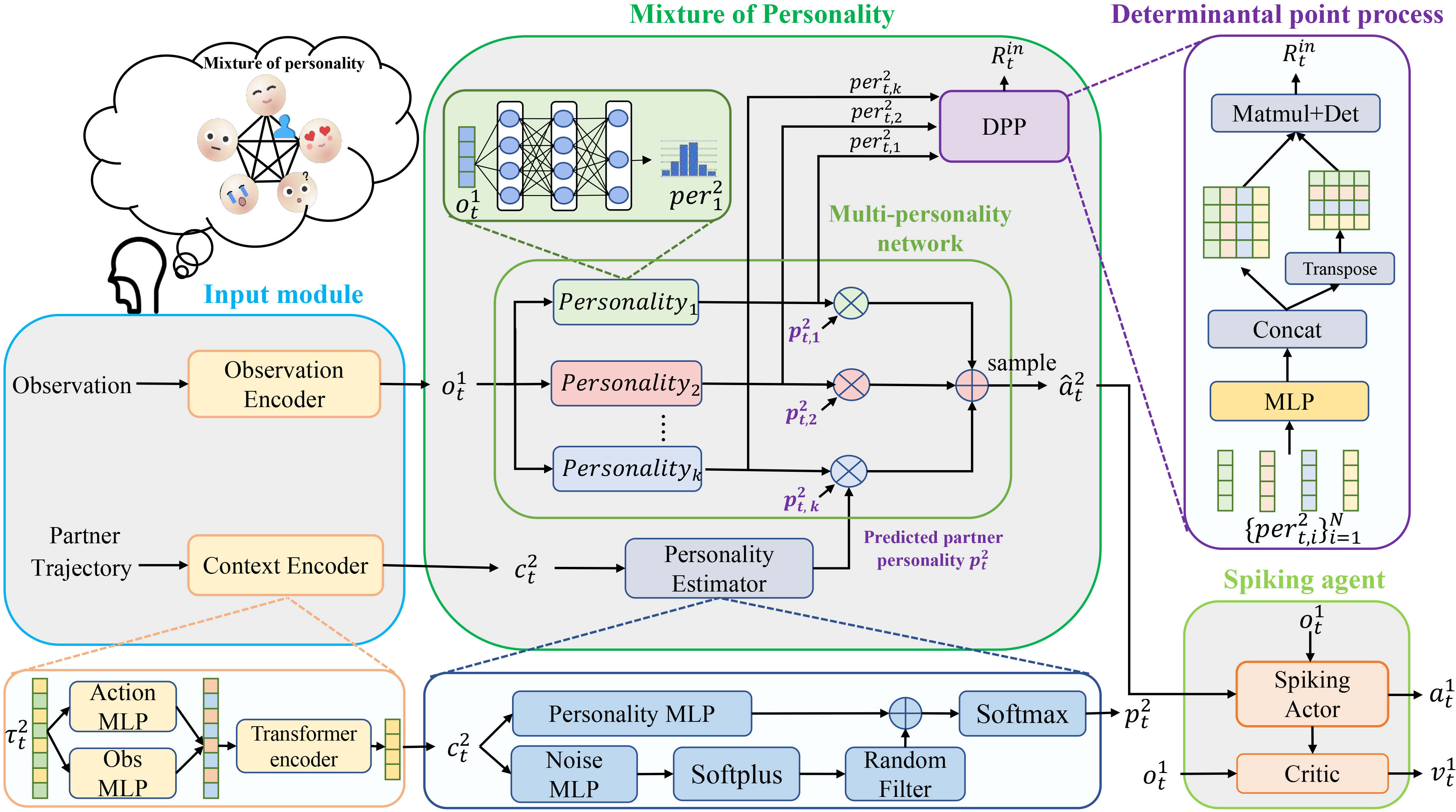}
  \caption{The detailed structure of MoP-SAN. The MoP-SAN consists of a SAN agent, a MoP model, and an input module that includes a context encoder and an observation encoder. The SAN PPO is used to simulate the ego agent with MoP. The MoP model is used to simulate the theory of mind process of our ego agent modeling the personality of the unseen partners. Our MoP model contains the personality estimator (PE) module, the multi-personality network, and the DPP module. 
  }
  \label{fig:mop}
\end{figure*}

In the last section, the cooperative MARL problem is defined. We present our algorithmic architecture and pipeline for the learning and generalization phases in this section. In this paper, we propose a robust framework for multi-agent collaboration. The left side of Figure~\ref{fig:overview} represents the two phases in our experiment, which will be discussed in the following section. The right side of Figure~\ref{fig:overview} shows the pipeline of our MoP-SAN in the zero-shot collaboration, and Figure~\ref{fig:mop} illustrates the detailed structure of our MoP-SAN.

As shown in Figure~\ref{fig:overview} and Figure~\ref{fig:mop}, our proposed framework includes a MoP model and a SAN model as the ego agent under the consideration of biological plausibility and energy efficiency. The MoP as partner mental model can understand the behavior of the partner and model the partner to estimate the personality of partner first, then instruct the action of the SAN agent. The SAN agent can have a better generalization ability of partner heterogeneity (zero-shot collaboration with diverse unseen partners) and cooperate with the unseen partner through the aid of the MoP model $m$. 

As shown in Figure~\ref{fig:overview}, we can divide our process into the learning and generalization phases, also called the training and testing process. We introduce a general framework that does not require additional expert-supervised data in the learning phase. In our current model, for simplicity, we assume that the observation encoder is an identity mapping, and the observation from environment is the input to the MoP. In order to self-supervise the training of the MoP model without additional expert data, we directly train MoP as a partner in the learning process for the sake of simplicity. 
 
On the one hand, the MoP model can act as a pool of many diverse agents to facilitate the learning of the SAN agent. On the other hand, the MoP model can also learn various personalities. In the generalization phase, we want to infer better and adapt to the unseen partner with a specific personality, so we need to discover as many base personalities in the personality space as possible during the learning process.

In the generalization phase, parameters in our framework are fixed. As shown in Figure~\ref{fig:overview}, when the SAN agent needs to cooperate with an unseen partner, the personality estimator (PE) determines the partner's personality first according to the historical context information of the unseen partner, and then the multi-personality network infers the current intention and action of the partner. Our goal is to maximize the total reward and entropy based on the historical information of the unseen partner.
In the following sections, our descriptions and formulas use the generalization phase as an example to describe our method.
The output of our MoP model is the input for the spiking policy of SAN $\rho_{\theta}^{1}$ and ${\theta^{1}}$ is the parameter for the policy network in SAN. ${\varphi}$ and ${\eta}$ are the parameter for the MoP model, and the joint policy can be written as follows:
\begin{equation}
\pi\left(\mathbf{a}_{t} \mid \mathbf{o}_{t}\right) = \rho_{\theta}^{1}\left(a_{t}^{1} \mid o_{t}^{1},\hat{a}_{t}^{2}\right) \rho_{\theta}^{2}\left(a_{t}^{2} \mid o_{t}^{2} \right)\textbf{,}
 \end{equation}
where $o_{t}^{i}$ is the observation of the $i$-th player, $\hat{a}_{t}^{2}$ denotes the predicted action distribution from our MoP model.

\subsection{The SAN model and context encoder}

The SAN model in our MoP-SAN refers to a SAN PPO agent, which makes its action based on the guided action of the MoP model to maximize the cooperation reward and entropy. The output action $a_t^{1}$ is sampled from the probability distribution over the action space of the spiking policy in the SAN model 
$\rho_{\theta}^{1}\left(a_{t}^{1} \mid o_{t}^{1},\hat{a}_{t}^{2}\right)$.  The SAN PPO agent includes a spiking actor and critic. The SAN model consists of leaky-integrate-and-fire (LIF) neurons, an abstraction of the Hodgkin-Huxley model. Non-differential membrane potential and refractory period are biologically plausible characteristics of the LIF neuron, which can simulate the neuronal dynamics. We define LIF neurons as follows:
\begin{equation}
    \tau \frac{d V_{t}}{d t}=-V_{t}+{I}_{t}
    \textbf{,}
\end{equation}

where $V_{t}$ represents the dynamic variable of membrane potential for time $t$, and $dt$ is the minimal simulation time slot. ${I}_{t}$ represents the integrated post-synaptic potential and $\tau$ is the integrative time period. With input ${I}_{t}$ within a period time of $\tau$. when $V_{t}$ is bigger than the firing threshold $V_{th}$, the neuron will be fired and generate a spike, and the membrane potential $V_{t}$ will be reset as the reset potential $V_{reset}$. The neuron will be mostly leaky when $V_{t}$ is smaller than the firing threshold. The detailed configuration of SAN is shown in our previous work \citep{zhang2022multi}.

The context encoder is the key to our good generalization and adaptation ability. We use the transformer model as our context encoder, and the input of our context encoder is the historical trajectories of the partner in a specific context size as context information. For context information, historical actions and observations have different dimensions. Therefore we introduce an action MLP network and obs MLP network to convert historical actions and observations into the same dimension, concatenating them in alternating order according to the order of time t in the trajectory $\tau$, similar to \citep{chen2021decision,meng2023offline}. 
\subsection{The MoP model}

The ToM ability of our MoP-SAN is delivered by our MoP model $m$, which consists of the multi-personality network, the PE module, and the DPP module. 

The multi-personality networks include $k$ different personality networks, each consisting of three-layer-MLP that represent a category of base personality strategies with a different policy. The input of our multi-personality network is the observation of the SAN agent and the output of $i$-th personality network $per_{t,i}^{2}$ is a action distribution corresponding to the respective basic personality under the same environmental observation.

The input of the PE module is the partner's context information $c_{t}^{2}$ which is the context embedding from historical trajectories of the partner by context encoder. 
In contrast to an entirely rational AI agent, the unseen partners are subject to some irrational factors that affect their decisions. Therefore our PE module consists of a personality multi-layer perceptron (MLP) represented by a trainable weight matrix $W_{p}$ and a Noise MLP represented by $W_{noise}$. The output of the Noise MLP is passed through a softplus function and a random filter, then added to the output of the personality MLP. The resulting sum is then passed through a softmax function to obtain an estimated personality profile $p_{t}^{2}$ for a unseen partner. The $e$ represents the PE function and the $R$ denotes a random filter function:
\begin{equation}
e(c_{t}^{2})=\textrm{Softmax}\left(c_{t}^{2} \cdot W_{p} + R (c_{t}^{2} \cdot W_{noise}) \right)\textbf{,}
\end{equation}

The output of the MoP model $\hat{a}_{t}^{2}$ is sampled from the probability distribution over the action space $m_{\varphi,\eta}\left(\hat{a}_{t}^{2} \mid o_{t}^{1}, c_{t}^{2}\right)$. 
The output of the PE module $p_{t}^{2}$ corresponds to the predicted partner personality,
and the output of $i$-th personality network ${per}_{t,i}^{2}$ represents the probability distribution over the action space of the $i$-th base personality in the current observation. 
$\eta$ is the parameter of the DPP in MoP and $\varphi$ is the parameter of the MoP model. The policy of our MoP can be defined as followings:
\begin{equation}
m_{\varphi,\eta}\left(\hat{a}_{t}^{2} \mid o_{t}^{1},c_{t}^{2}\right)=\sum_{i=1}^{n}p_{t,i}^{2} \cdot per_{t,i}^{2}\textbf{,}
\end{equation}

Instead of a sparsely-activated model that chooses different branches for different tasks, our MoP method integrates the output of all the base personalities rather than selecting a base personality each time. Therefore, 
the output of the PE module, the predicted personality of the partner, is not a discrete one-hot vector but a floating-point vector that sums to one. 

Our MoP can model partners and infer the personalities of other partners that can help any RL agents to enhance their generalization ability and adaptability so that the agent can be applied to many zero-shot collaboration scenarios. 

\subsection{The DPP module in the MoP}

In this section, we introduce the DPP first and present the DPP in our proposed MoP-SAN. DPP \citep{kulesza2012determinantal} is an efficient probabilistic model proposed in random matrix theory and has been widely used in many application fields of machine learning \citep{gong2014diverse,perez2021modelling,parker2020effective}, such as recommendation systems \citep{chen2018fast}, and video summarization \citep{gong2014diverse}. The high-performing model DPP can translate complex probability computations into simple determinant calculations and then use the kernel matrix's determinant to calculate the probability of each subgroup. 
Recent studies, such as \citep{dai2022diversity,yang2020multi}, have incorporated the DPP model into reinforcement learning (RL) approaches. \citep{dai2022diversity} utilized DPP models to introduce intrinsic rewards and enhance the exploration of RL methods. Meanwhile, \citep{yang2020multi} used DPP to enhance existing RL algorithms by encouraging diversity among agents in RL evolutionary algorithms.

In the learning process, the multi-personality network can be considered to have various personalities. Each personality network can be regarded as a distinct base personality. 
Measuring the diversity among the multiple base personalities is crucial for constructing a diverse set of base personalities in the personality space.
To effectively explore the range of personalities in task space, we integrate a diversity-promoting DPP module to regularize these base personalities in our MoP-SAN. This ensures efficient exploration and optimization of the diverse set of personalities, improving the overall performance of our MoP-SAN.

We can measure the diversity of the personalities and select the subset of diverse personalities through the diversity constraints as an intrinsic reward imposed by the DPP module. $Y$ denotes the set containing many personalities, and $y$ refers to a subset of $Y$ including $k$ personalities that can maximize the diversity. 
Since these personality networks share the same observation input, and the output of a specific personality network $per_{t,i}^{2}$ is an action distribution, therefore the difference between base personalities can be measured by the action distribution over the action space. 
We denote the kernel matrix of $y$ as $L_{y}$. The determinant value of $L_{y}$ can represent the diversity of the personality set $y$. To construct the set $y$, we need to select $k$ personalities in the personality space for maximizing the determinant value of the kernel matrix of $y$.
The personality set $y$ can be regarded as a set of base personalities that maximizes diversity in the personality space. 
\begin{equation}{y}^{*}=\arg \max _{{y}} P(Y={y})=\arg \max _{{y}} \operatorname{det}\left({L}_{{y}}\right)\textbf{,}\end{equation} 
Since the matrix $L_{{y}}$ is positive semi-definite, there exists matrix $B_{t}$ at every time step $t$ such that:
\begin{equation}L_{{y}}=B_{t}B_{t}^{T}\textbf{,}\end{equation}
$B_{t}$ and the intrinsic reward $r^{\textrm{dpp}}_{t}$ can be defined as follows, and $k$ is the number of personalities: 
\begin{equation}B_{t}=\left[\upsilon_{\eta}\left(per_{t,1}^{2}\right), \upsilon_{\eta}\left(per_{t,2}^{2}\right), \upsilon_{\eta}\left(per_{t,3}^{2}\right),\ldots,\upsilon_{\eta}\left(per_{t,k}^{2}\right)\right]^{T}\textbf{,}\end{equation}
\begin{equation} r^{\textrm{dpp}}_{t}\left(per_{t,1}^{2},per_{t,2}^{2} \ldots per_{t,k}^{2};\eta\right) = \log \textrm{det}\left({B_{t} } {B_{t}}^{T}\right )\textbf{.}
\end{equation}
where $\upsilon_{\eta}$ represents the feature vector parameterized by the parameters $\eta$.

We endeavor to build some unique personality vectors as base personalities for our multi-personality network, which can combine the entire personality space. Therefore, our MoP model with our proposed DPP module can enable rapid adaptation and generalization to any unseen partners in the collaboration task.
\subsection{The SAN learning}
The policy parameter of the SAN agent $\theta^{1}$ and the MoP model parameter $(\varphi,\eta)$ are iteratively optimized in our method. The overall optimization objective is to maximize the cumulative discounted return, which depends on the MoP model $m_{\varphi, \eta}\left(a_{t}^{2} \mid o_{t}^{2},c_{t}^{2}\right)$ and the spiking policy of the SAN agent $\rho_{\theta}^{1}\left(a_{t}^{1} \mid o_{t}^{1},a_{t}^{2}\right)$:
\begin{gather}
\begin{aligned}
{\theta^{1}}^{*},\varphi^{*},\eta^{*} =\max _{\theta,\varphi,\eta}\sum_{t=0}^{\infty} \mathbb{E}_{a_{t}^{1},a_{t}^{2}}\left[ \gamma^{t}\left(r\left(\mathbf{o}_{t}, \mathbf{a}_t\right) 
+\alpha\overline{\mathcal{H}}\left(\pi\left(\mathbf{a}_t \mid \mathbf{o}_{t}\right)\right)\right)\right] 
\end{aligned}\textbf{,}
\end{gather}
The goal of the SAN agent is to maximize the extrinsic reward $r^{\textrm{ex}}_{t}$ by collaborating with partners. We can calculate the gradient of the SAN as follows:
\begin{equation}
    \begin{aligned}
        \nabla_{\theta} J\left(\rho_{\theta}^{1}\right)&=
\mathbb{E}_{a_{t}^{1},a_{t}^{2} }\Bigl[ \nabla _ { \theta  } \operatorname {log} ( \rho_{\theta}^{1}( a_{t}^{1}|o_{t}^{1},a_{t}^{2} ) )  \bigl( G^{ex}(\mathbf{o}_{t}, \mathbf{a}_t)\\& -b_{1}\left(o_{t}^{1}, \mathbf{a}_t\right)-\alpha \log (\rho_{\theta}^{1}(a_{t}^{1} \mid o_{t}^{1},a_{t}^{2}))\bigr)\Bigr] \\
&  a_{t}^{1} \sim \rho_{\theta}^{1}\left(a_{t}^{1} \mid o_{t}^{1},a_{t}^{2}\right),a_{t}^{2} \sim m_{{\varphi,\eta}}\left(a_{t}^{2} \mid o_{t}^{2},c_{t}^{2}\right)
\end{aligned}\textbf{,}
\end{equation}
where the $b_{1}$ is the baseline function. We can estimate the baseline function $b_{1}$ by the expected return of all possible actions, as shown in follows:
\begin{equation}
b_{1}\left(o_{t}^{1}, {a}_t^{1}\right)=\sum_{a_{t}^{1} \in \mathcal{A}} \rho_{\theta}^{1}\left(a_{t}^{1} \mid o_{t}^{1},a_{t}^{2}\right) G^{ex}\left(\mathbf{o}_{t},\mathbf{a}_t\right)\textbf{.}
\end{equation}
where $G^{\textrm{ex}}\left(\mathbf{o}_{t}, \mathbf{a}_t\right)$ denotes the discounted extrinsic returns for SAN and MoP. Here we use the game score as the extrinsic reward $r^{\textrm{ex}}_{t}$.

\subsection{The MoP learning}

We introduce the DPP constraint into our work, similar to recent work \citep{dai2022diversity}, by treating the DPP diversity measurement as the intrinsic reward. We adopt a bi-level optimization framework \citep{dai2022diversity} for the MoP model and its DPP module to maximize the intrinsic reward and extrinsic reward.

Our objective can be defined as follows:
\begin{equation}
\max _{\eta} J^{\textrm{ex}}\left(\varphi^{\prime} ,\eta\right)
\text { s.t. } \varphi^{\prime}=\underset{\varphi}{ argmax } J^{\textrm{mix}}(\varphi,\eta)\textbf{,}
\end{equation}
For this optimization problem, we can treat it as a Stackelberg game. We use the DPP reward as the intrinsic reward. The mixture rewards are the sum of intrinsic and extrinsic rewards. The mixture reward can be written as follows: 
\begin{equation}
r^{\textrm{mix}}_{t} = r^{\textrm{ex}}_{t}+\beta r^{\textrm{ {dpp}}}_{t}\left(a_{1},a_{2} \ldots a_{k};\eta\right)\textbf{,}
\end{equation}
where $\beta$ is the weight coefficient of the intrinsic reward. $r^{\textrm{ex}}_{t}$ is the standard reward from the environment where the SAN agent makes actions $a_{t}^{1}$, and MoP makes $a_{t}^{2}$ in the environmental state $s_{t}$ at the time step $\mathit{t}$, and $r^{\textrm{dpp}}_{t}$ is the DPP constraint diversity reward for the partner. The gradient $\nabla_{\varphi} J^{\textrm{mix}}$ can be calculated as follows:
\begin{equation}
    \begin{aligned}
 \nabla_{\varphi} J^{\textrm{mix}} &= \alpha \cdot \nabla_{\varphi} \log m_{\varphi,\eta}\left(a_{t}^{2} \mid o_{t}^{2},c_{t}^{2}\right)\Bigl( G^{\textrm{mix}}\left(\mathbf{o}_{t}, \mathbf{a}_t\right)\\
 &-b_{2}\left(o_{t}^{2}, \mathbf{a_{t}}\right)-\alpha \log \left(m_{\varphi,\eta}\left(a_{t}^{2} \mid o_{t}^{2},c_{t}^{2}\right)\right) \Bigr)
    \end{aligned}\textbf{,}
\end{equation}
where $G^{\textrm{mix}}\left(\mathbf{o}_{t}, \mathbf{a}_t\right)$ denotes the discounted mixture returns for our MoP-SAN.
The gradient $\nabla_{\eta} J^{\textrm{ex}}$ can be calculated by using the chain rule:
\begin{equation}
    \nabla_{\eta} J^{\textrm{ex}}=\nabla_{\varphi^{\prime}} J^{\textrm{ex}} \nabla_{\eta} \varphi^{\prime}\textbf{,}
\end{equation}
with
\begin{equation}
    \begin{aligned}
\nabla_{\eta} \varphi^{\prime} &=\nabla_{\eta} \alpha G^{\operatorname{mix}}\left(\mathbf{o}_{t}, \mathbf{a}_t\right) \nabla_{\varphi} \log m_{\varphi,\eta}\left(a_{t}^{2}|o_{t}^{2},c_{t}^{2}\right) \\
&=\alpha \beta \sum_{l=0}^{\infty} \gamma^{l} \nabla_{\eta} R_{\eta, t+l}^{\text {d}} \nabla_{\varphi} \log m_{\varphi,\eta}\left(a_{t}^{2}|o_{t}^{2},c_{t}^{2}\right)
\end{aligned}\textbf{,}
\end{equation}
We can use importance sampling to improve the sample efficiency of the algorithm:
\begin{equation}
    \nabla_{\varphi^{\prime}} J^{\textrm{ex}}=\nabla_{\varphi^{\prime}}\left(\frac{m_{\varphi ^{\prime},\eta}\left(a_{t}^{2}|o_{t}^{2},c_{t}^{2}\right)}{m_{\varphi,\eta}\left(a_{t}^{2}|o_{t}^{2},c_{t}^{2}\right)}\right) G^{\operatorname{mix}}\left(\mathbf{o}_{t}, \mathbf{a}_t\right)\textbf{,}
\end{equation}
\begin{equation}
\begin{aligned}
    \nabla_{\eta} J^{\textrm{ex}}&=\nabla_{\varphi^{\prime}} J^{\textrm{ex}} \nabla_{\eta} \varphi^{\prime} \\
    &=\nabla_{\varphi^{\prime}}\left(\frac{m_{\varphi ^{\prime},\eta}\left(a_{t}^{2}|o_{t}^{2},c_{t}^{2}\right)}{m_{\varphi,\eta}\left(a_{t}^{2}|o_{t}^{2},c_{t}^{2}\right)}\right) G^{\textrm{mix}}\left(\mathbf{o}_{t}, \mathbf{a}_t\right)\alpha \beta\cdot  \\
    & \sum_{l=0}^{\infty} \gamma^{l} \nabla_{\eta} R_{\eta, t+l}^{\textrm{dpp}} \nabla_{\varphi} \log m_{\varphi,\eta}\left(a_{t}^{2}|o_{t}^{2},c_{t}^{2}\right)
\end{aligned}\textbf{.}
\end{equation}
Hence, the iterative learning of policy parameters in the SAN and MoP model finally converges the whole system to support next-step MARL tasks. 
\section{Experimental results}

\subsection{Environmental setttings}

Our experimental environment is Overcooked \citep{carroll2019utility}, a primary human-AI zero-shot collaboration benchmark. Similar to previous works 
\citep{carroll2019utility,shih2021critical,shih2022conditional}, we have conducted experiments on the ``simple" map based on PantheonRL \citep{sarkar2022pantheonrl}, a pytorch framework for human-AI collaboration. In this environment, two players cooperate to complete the cooking task, i.e., making as many onion soups as possible for winning a higher reward in a limited time. The players can choose one of six actions and execute simultaneously, including up, down, left and right, empty operation, or interaction. 

It is necessary to follow a specific order when making onion soup. The player must put three onions in the pot and cook them for 20 steps. Then player pours the onion soup from the pot onto the plate and serves the dish to the designated position. After this process, the player can get certain rewards (20). A player can not complete this task alone on the challenging task. Only through good cooperation can the players achieve high scores, which requires the ability to infer the personality of the partner first and predict the actions of the partner. 
 
\begin{figure*}[htb]
  \centering
\includegraphics[width=12cm]{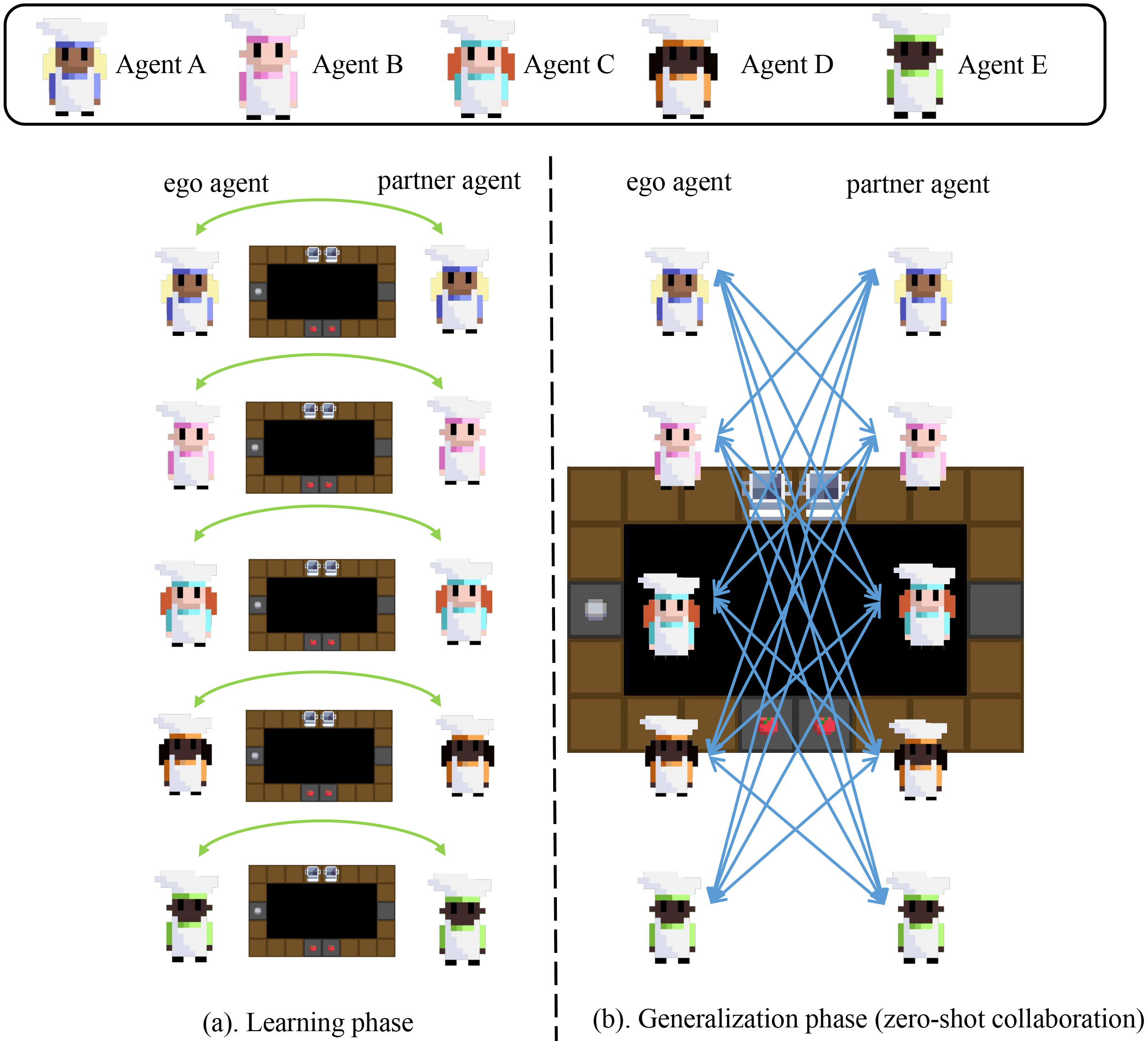}
  \caption{The experimental setting detail for our MoP-SAN in learning and generalization phases (zero-shot collaboration). Agent A-E corresponds to the agent with different seeds whose name is A-E. (a). In the learning phase, the ego agent and specific partner agent in a pair collaborate for this task and are trained by iterative optimization. The ego agent and partner agent in a pair have the same name. There are five agent pairs in the learning phase: (A, A), (B, B), (C, C), (D, D), and (E, E). (b). In the generalization phase, the ego agent needs to collaborate with all unseen partner agents in a zero-shot manner. For example, the ego agent A will cooperate with another unseen partner agent with a different name (B, C, D, or E) for the zero-shot collaboration test.}
  \label{fig:detail}
\end{figure*}

\subsection{Configurations of our baselines and our MoP-SAN}

There are several baseline methods. One method is the standard DNN PPO baseline \citep{schulman2017proximal}, an important MARL method with excellent performance in many scenarios. In this method, both the ego and the partner agent are homogeneous PPO agents, and this way is also called self-play \citep{silver2018general} in RL. 

Another important baseline is the SAN PPO baseline. Since the ego agent in our method is also the SAN PPO, we refer to the SAN PPO baseline as the SAN baseline in the following experimental description. It is worth mentioning that we first introduce the SAN version of PPO into the multi-agent cooperation task Overcooked. For the SAN baseline, in our cooperation environment, the ego agent is the SAN PPO, and the partner is the standard PPO.

The experimental details of our setting are shown in Fig. \ref{fig:detail}. As shown in Figure~\ref{fig:overview} and Figure~\ref{fig:detail}, the SAN agent and MoP in one pair have the same name and are trained together by iterative optimization in the learning phase for our MoP-SAN. For example, our SAN A as the ego agent and MoP A as the partner will cooperate in the learning phase for a good score. In the generalization phase, SAN and MoP with the same name will be combined into MoP-SAN as the ego agent. We will evaluate the generalization of our proposed MoP-SAN model by cooperating with different unseen partners, which means the ego and partner agent in one pair have different names.

Our training experiment is run for half a million steps, and the generalization experiment (zero-shot collaboration) is conducted for several games to take the average score during the generalization phase in all our experiments. The personality number is 12, and the context size is 5. For the context encoder in our MoP-SAN, if the length of historical trajectories of the partner is less than the context size, we will pad 0. We use a single-layer transformer with two heads as a context encoder whose inner dimension is 256 and the dimension for q,k,v is 64. For the part of padding 0, we mask it in the transformer. Our MoP-SAN model uses an actor-critic framework, and the actor is based on SAN, similar to previous work \citep{zhang2022multi}. The actor network is (64, tanh, 64, tanh, 6); the critic network is (64, tanh, 64, tanh, 1). We sample action from categorical distribution for all methods. In these methods, we use the adam optimizer, and the learning rate is 0.0003. The reward discount factor is $\gamma = 0.99$, and the batch size is 64. The weight coefficient of the intrinsic reward $\beta$ is 0.5, and the maximum length of the replay buffer is 2048. We use gradient clipping to prevent exploding and vanishing gradients. 

\subsection{Stronger generalization ability of MoP-SAN}

\begin{figure}[htb]
\centering
\includegraphics[width=16cm]{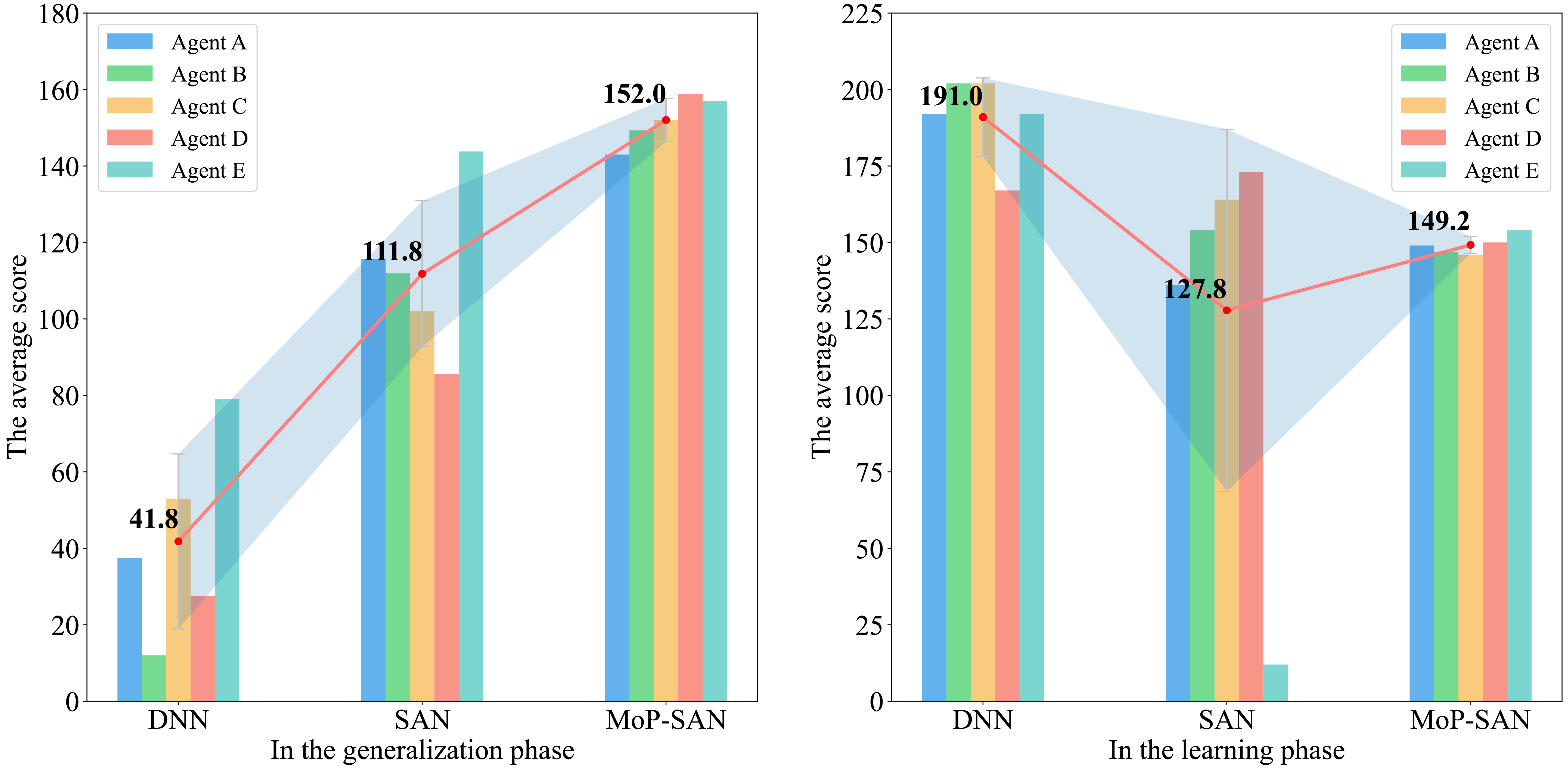}
\caption{Score comparison between baseline and MoP-SAN models in the generalization and learning phases. The left figure shows that our MoP-SAN outperforms other baselines in terms of generalization performance, with strong generalization ability to complete cooperative tasks with unseen partners. The right figure shows that our MoP-SAN improves the poor performance and large variance of SAN in the learning phase. Agents A-E denote different agents with different random seeds.}
\label{fig_traintestall}
\end{figure}

Fig. \ref{fig_traintestall} is a histogram representing the generalization and learning scores obtained by three methods in the Overcooked task. 
The line chart in the histogram shows the trend of the average score for the different methods. The red dot indicates the average score of all corresponding agents, and the shaded area represents the standard deviation of the corresponding results for the three methods. 

The average score for the method in the left diagram is the average score of all generalization tests with unseen partners. As shown in Fig. \ref{fig:detail}, the average score for our MoP-SAN method in A is 142, which means that the average for four unseen tests (A-B, A-C, A-D, A-E) is 142. The average score for our method is 142.25 means that the average for twenty unseen tests (A-B, A-C, A-D, A-E, B-A, B-C, B-D, B-E, C-A$\dots$) is 142.25. Fig. \ref{fig_methodsall} shows the detailed score for all generalization tests with unseen partners. The detailed score in the learning and generalization phase for each pair can be found in the supplementary material.

Fig. \ref{fig_traintestall} indicates that 
our proposed MoP-SAN model outperforms all baselines for unseen partners during the zero-shot collaboration, showing a more robust and stable ability for cooperation. 
What needs to be further emphasized is that our MoP-SAN method not only significantly outperforms the SAN baseline but also the DNN baseline in the generalization test, which strongly demonstrates the powerful generalization ability for partner heterogeneity of our method in zero-shot collaboration.

The average score in learning phase can be found in the right diagram of the Fig. \ref{fig_traintestall}. Although our MoP-SAN method primarily focuses on zero-shot generalization test without any prior knowledge of partners, the scores during the learning phase can still reflect the collaborative performance with the specific partner. Our MoP-SAN has better learning scores and minor variance compared to the SAN baseline in the learning phase. 

\subsection{Significantly better zero-shot collaborative performance of MoP-SAN}
\begin{figure}[htb]
\centering
\includegraphics[width=16cm]
{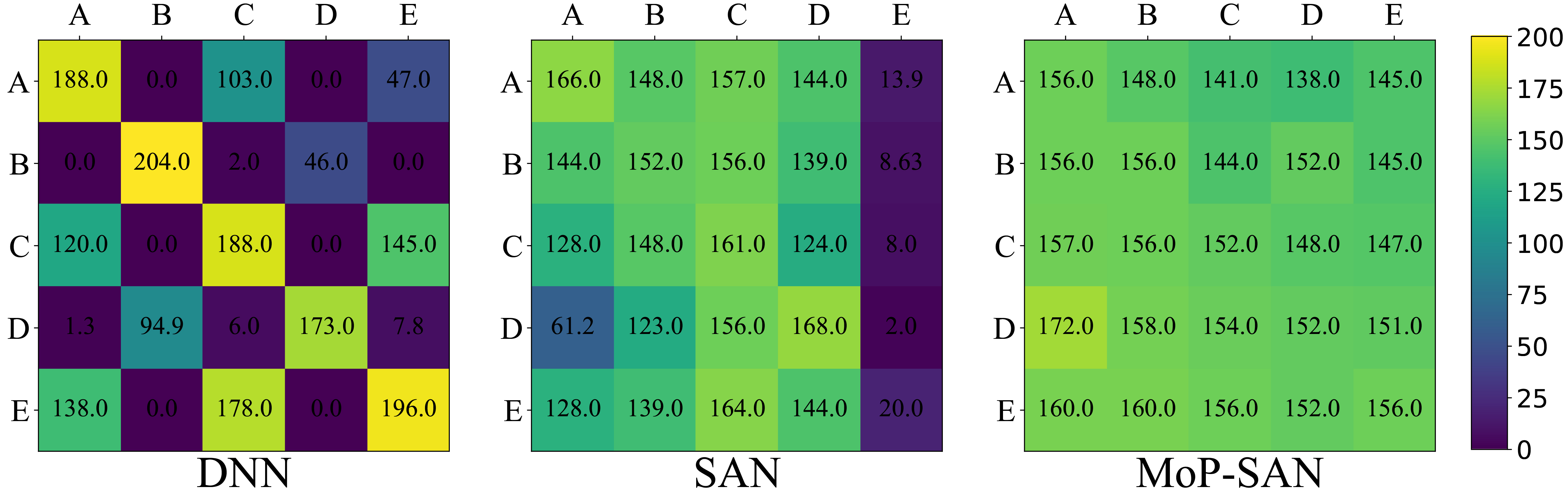}
\caption{The color temperature diagram shows the detailed generalization score for the baseline methods and our MoP-SAN. The difference in colors demonstrates the difference in scores. 
Compared with the DNN and SAN baseline, our proposed MoP-SAN has more satisfactory results for a better score and smaller variance. 
}
\label{fig_methodsall}
\end{figure}
Our experimental results in the zero-shot collaboration test reflect the generalization ability of partner heterogeneity for different methods. Fig. \ref{fig_methodsall} is the color temperature map showing the specific experimental data in the generalization test for all three methods. The color temperature maps in Fig. \ref{fig_methodsall} correspond to the DNN baseline, the SAN baseline, and our MoP-SAN model, respectively. The row represents the ego agent, and the column represents the partner. For example, the score in (the first row, the third column) for our MoP-SAN, represents the zero-shot collaboration score between MoP-SAN A and unseen partner C. The scores on the diagonal represent the scores achieved by the corresponding pairs during the learning phase, which are not included in the zero-shot collaboration score data of the generalization phase. We can see that the more obvious the color difference is, the more significant the variance of this method. 

As shown in Figure~\ref{fig_methodsall}, our multi-scale biological plausibility MoP-SAN achieved significantly better scores and smaller variance than the other baselines for most pairs 
in the zero-shot generalization test with low energy consumption, achieving good generalization results with unseen partners of different styles. Although DNN achieves high scores in some generalization test experiments, its variance is large, and the average score is low. Moreover, the SAN baseline has a better average score and smaller variance than the DNN baseline. 
These results demonstrate that our MoP model can complete partner modeling and help the SAN agent have a higher collaborative score with better generalization ability. 

\begin{figure}[htb]
\centering
\includegraphics[width=16cm]
{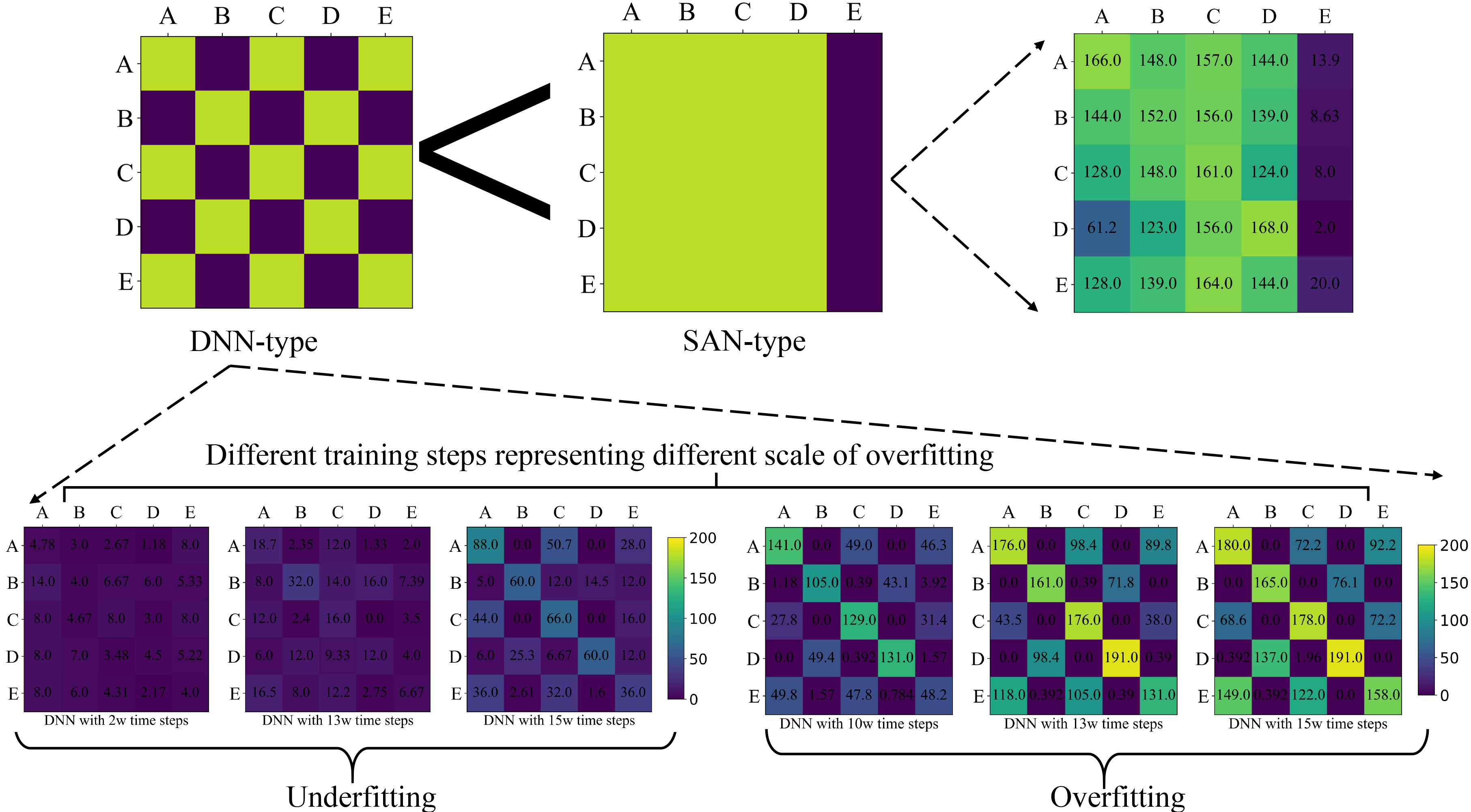}
\caption{The diagram depicts the detailed generalization analysis experiment of DNN and SAN, showing the generalization test results of the DNN under different training steps, which represent different scales of overfitting. As the number of training steps increases, the generalization performance of DNN gradually improves. The generalization test results for DNN exhibit a similar pattern of DNN-type, while the results for SAN also exhibit a similar pattern of SAN-type. By comparing these two patterns, we can see that SAN has better generalization ability and robustness.}
\label{fig_analysisall}
\end{figure}
The question of why SAN can achieve better generalization results than DNN has caught our attention. In order to further verify whether the poor generalization test performance of DNN was due to overfitting, we conducted a series of analysis experiments on DNN. We saved the checkpoints of DNN's learning process from underfitting to ``overfitting" and performed unseen partner generalization tests. These results indicate that as the number of training steps increases, the generalization performance of DNN gradually improves. We have discovered a similar pattern in these test results and named it DNN-type. 

Similarly, in the generalization test results of SAN, we also discovered a similar pattern which we named SAN-type. Compared to DNN-type, SAN-type exhibits stronger generalization and cooperation abilities in unseen partner generalization scenarios. These results represent that ``overfitting" was not the main cause of the poor generalization test performance of DNN. We believe that the reason why DNN performs worse than SAN in the generalization test with unseen partners is that SAN has better noise resistance and robustness. In cooperative reinforcement learning, the generalization test with unseen partners can be regarded as a noise perturbation test, and therefore SAN performs better than DNN in our generalization experiment. 
\subsection{Larger personality size contributes better cooperative performance}
Furthermore, we conduct some ablation experiments to confirm the effectiveness of different modules and parameters in our MoP-SAN.
\begin{table*}[htb]
\begin{center}
\caption{The mean score of different number of personalities in our method.}
\vspace{5pt}
\begin{tabular}{ccccccc@{}}
\toprule
\textbf{{Agents}}  &  A &  B    & C    & D    & E    & avg      \\ \midrule
\textbf{ours w$/$personality 6}  & 0.2 & 0.4 & 0.4 & 0 & 1.6 & 0.52 ($\pm$0.63) \\
\textbf{ours w$/$personality 8}    & 1.8 & 123 & 0 & 0.4 & 0 & 25.04 ($\pm$54.77)  \\
\textbf{ours w$/$personality 10}    & 7.6 & 151 & 1.6 & \textbf{157} & 114 & 86.24 ($\pm$76.35)  \\
\textbf{ours w$/$personality 12}  & \textbf{149} & \textbf{154} & \textbf{150} & 146 & \textbf{146} & \textbf{149 ($\pm$3.32)}  \\
\bottomrule
\end{tabular}
\label{tab:robust_mean_train_expertnum}
\end{center}
\end{table*}
The experimental results in Table \ref{tab:robust_mean_train_expertnum} show that as the number of personalities increases, the learning ability of our MoP-SAN model gradually improves and the variance gradually gets smaller. These results also show that diverse personalities play an essential role in the multi-agent cooperation task. 

From Table \ref{tab:robust_mean_train_expertnum}, we can see that some pairs have very poor cooperation scores when the number of base personalities is small. This may be because these base personalities can not be combined to express all the dimensions of the personality of the partners. As the number of base personalities increases, the expression ability of the existing base personalities for personality of the current partner grows, resulting in better performance.

The personality theory in cognitive psychology suggests that breaking down personality into finer-grained traits is an excellent way to improve predicting and explaining human behavior. Our experimental results further validate this point. By using a larger personality number, we obtain more precise personality delineation, which can better predict the personality of the partner and cooperate more efficiently with partners to achieve higher scores.
\subsection{Richer context information contributes better personality prediction}
\begin{table}[htb]
\begin{center}
\caption{The mean score of different number of context size in our method.}
\vspace{5pt}
\begin{tabular}{ccccccccc@{}}
\toprule
\textbf{{Agents}}  &  A &  B    & C    & D    & E    & avg      \\ 
\midrule
\textbf{ours w$/$context 0}  & 4.4 & 2.8 & 2.6 & 1 & 1.2 & 2.4 ($\pm$1.38) \\
\textbf{ours w$/$context 1}  & 11.6 & 0.2 & 1 & 17.2 & 85.2 & 23.04 ($\pm$35.48) \\
\textbf{ours w$/$context 3}    & 136 & 148 & 143 & 140 & 140 & 141.4 ($\pm$4.45)  \\
\textbf{ours w$/$context 5}  & \textbf{149} & \textbf{154} & \textbf{150} & \textbf{146} & \textbf{146} & \textbf{149 ($\pm$3.32)}  \\
\bottomrule
\end{tabular}
\label{tab:robust_mean_train_contextsize}
\end{center}
\end{table}
Table \ref{tab:robust_mean_train_contextsize} indicates that as the context information of the partner increases, the score of our MoP-SAN in the learning phase gets better and better, which shows that partner information is crucial for our MoP-SAN model in the cooperation task. The result is the worst when there is no partner information at all. This is because partner information serves as input for the PE module to predict the personality of partner. Without such information, the personality prediction is random, leading to inefficient collaboration between ego and partner agents when completing tasks such as making onion soup. Limited partner information may make the personality prediction inaccurate, which is detrimental to the collaboration score.

These results in Table \ref{tab:robust_mean_train_contextsize} also indicate that the existence of partner context information is the key to our ability to solve this task. We find that the existence of partner information achieves better results in the learning phase and gets better generalization results in the zero-shot collaboration generalization experiment.
\subsection{Personality diversity controlled by DPP}
\begin{figure}[htbp]
\centering
\begin{minipage}[t]{0.48\textwidth}
\centering
\includegraphics[height=7cm] 
 {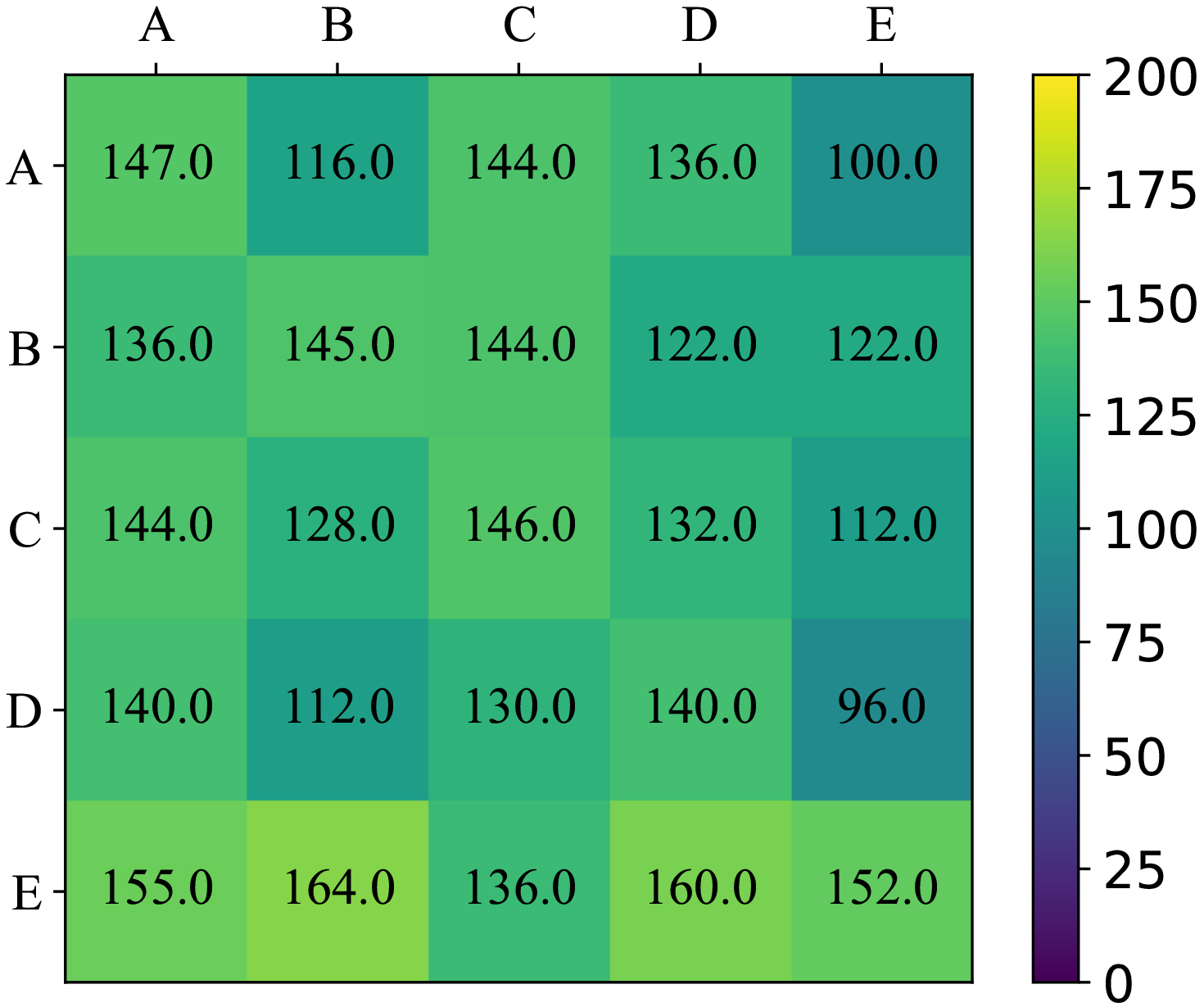}
\end{minipage}
\begin{minipage}[t]{0.48\textwidth}
\centering
\includegraphics[height=7cm]{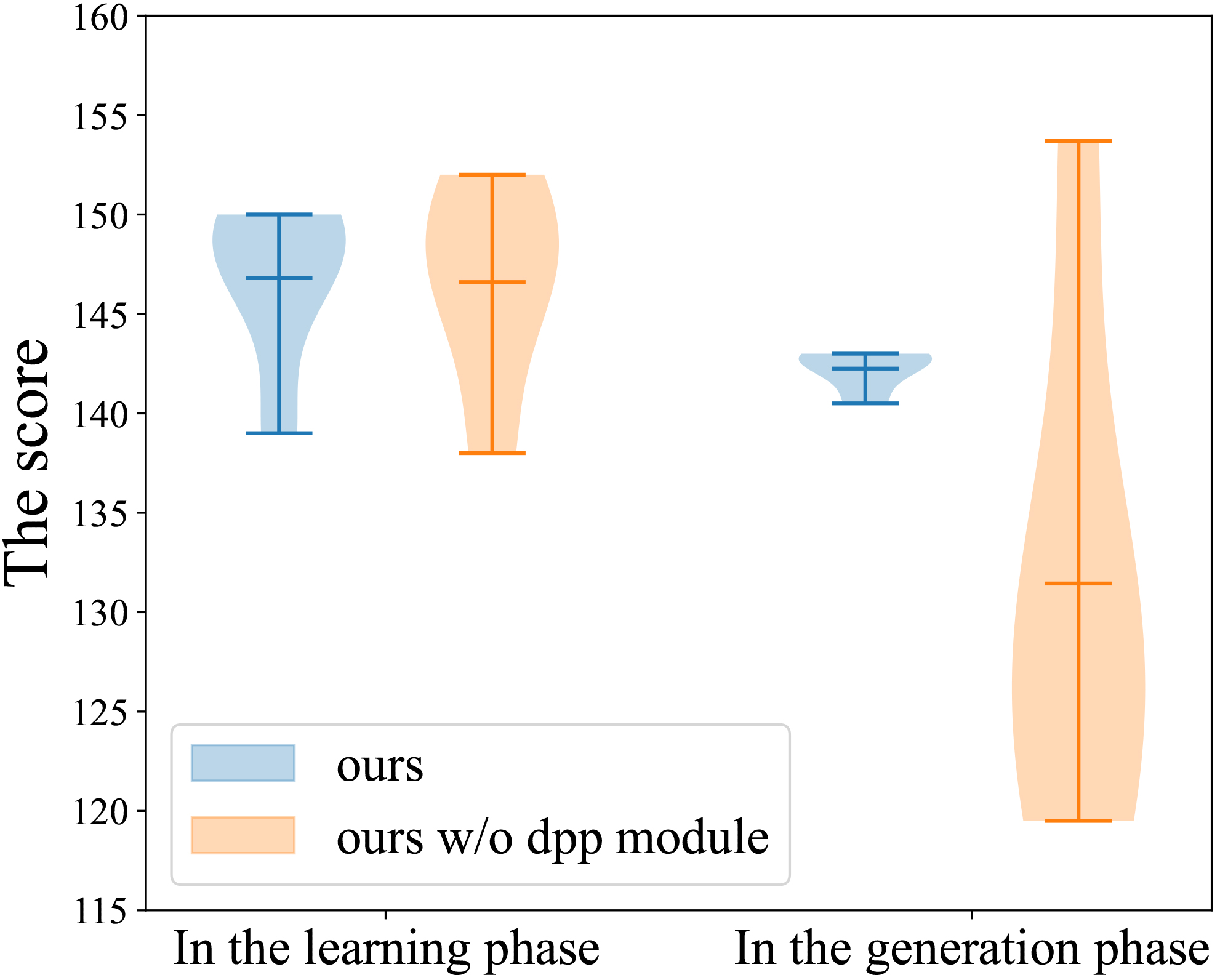}
\end{minipage}
\caption{The left color temperature diagram demonstrates the detailed generalization scores for our method w/o the DPP module. 
The right violin plot demonstrates the visual comparison of the scores in the ablation experiments on the DPP module, where violin plots are presented for our method w/o the DPP module and our method for the learning and generalization phases.}
\label{fig_dppall}
\end{figure}
The results in the ablation experiment of DPP demonstrate the effectiveness of the DPP module, which can achieve better results in the generalization experiments. We further analyze the results of the ablation experiment through the color temperature map and violin plot in Fig. \ref{fig_dppall}. We show the maximum, minimum, and average lines in the violin plot, and the shade means the data distribution whose size represents the variance of the corresponding method. As shown in the right violin plot of Fig. \ref{fig_dppall}, our method is much better than our method w/o DPP at the generalization test, and our MoP-SAN has a smaller variance than our MoP-SAN w/o DPP. The color temperature plot of our MoP-SAN is shown in Fig. \ref{fig_methodsall} as the third plot c. The comparison between the left color diagram in Fig. \ref{fig_dppall} with plot c in Fig. \ref{fig_methodsall} indicates that our MoP-SAN model has better generalization performance and minor variance owing to the DPP module. 

This result indicates that with the same size of personality number, the addition of DPP can constrain the base personalities in MoP, which allows these base personalities to cover as much personality space as possible. This complete coverage leads to a more robust PE module that can more accurately predict the personality of unseen partner, achieving in better scores.
\section{Conclusion}
In this paper, we focus on strengthening the conventional actor network by incorporating multi-scale biological inspirations, including the local scale neuronal dynamics with spike encoding and global scale personality theory with the spirit of the theory of mind. Our proposed mixture of personality improved spiking actor network (MoP-SAN) algorithm can remarkably improve the generalization and adaptability in the MARL cooperation scenarios, under a surprisingly low energy consumption. 

Our MoP-SAN is then verified by experiments, which shows that the two-step process in personality theory is very crucial for predicting the unseen partner's actions. The MoP improved SAN shows a more satisfactory learning ability and generalization performance compared with SAN and DNN baseline. To the best of our knowledge, we are the first to apply SAN and MoP in the MARL cooperation task. This integrative success has given us more confidence about borrowing more inspirations from neuroscience and cognitive psychology in the future for designing new-generation MARL algorithms.

\section*{Acknowledgment}
The authors would like to thank Yali Du, Dengpeng Xing, Zheng Tian, and Duzhen Zhang for their previous assistance with the valuable discussions. 

\section*{Author Contributions}
B.X, J.S, T.Z, X.L. gave the idea; X.L., Z.N. made the experiments and the result analyses; X.L., J.R, L.M were involved in problem definition. They wrote the paper together and approved the submitted version. 

\section*{Conflict of Interest Statement}
The authors declare that the research was conducted in the absence of any commercial or financial relationships that could be construed as a potential conflict of interest.

\section*{Funding}
This work was funded by the Strategic Priority Research Program of the Chinese Academy of Sciences (XDA27010404, XDB32070100), the Shanghai Municipal Science and Technology Major Project (2021SHZDZX), and the Youth Innovation Promotion Association CAS.

\bibliographystyle{Frontiers-Harvard} 
\bibliography{frontiers}

\end{document}


\onecolumn
\firstpage{1}

\title {{\helveticaitalic{Supplementary Material}}}
\maketitle

\section{Supplementary Tables and Figures}

\subsection{Figures}
\begin{figure}[htbp]
\begin{center}
\includegraphics[width=15cm]{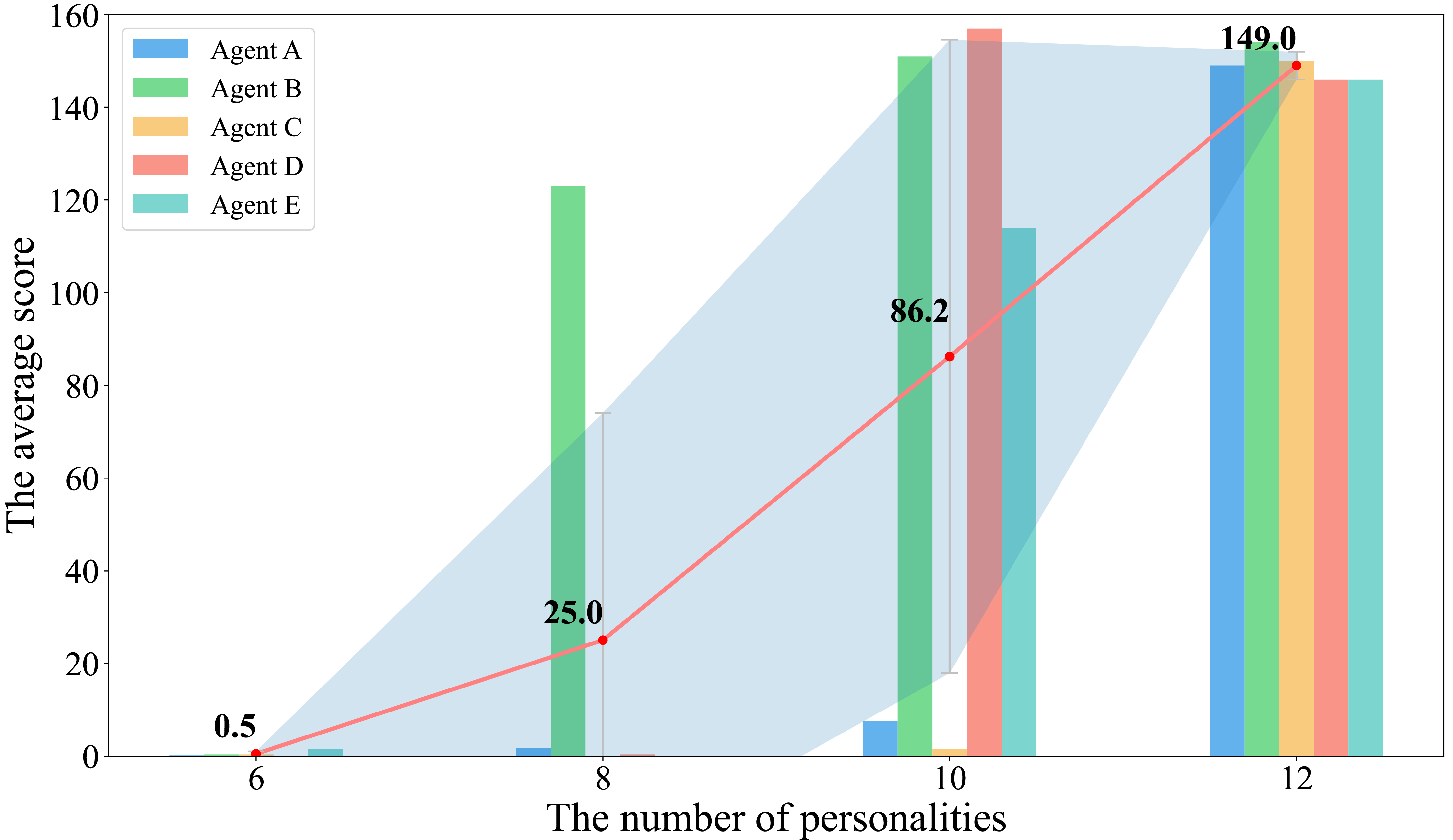}
\end{center}
\caption{This diagram indicates the detailed scores for different personality numbers (6, 8, 10, 12). A-E denotes different agent with different random seeds. In our method, the score drastically improves as the number of personalities increases.}
\label{fig_expertsall}
\end{figure}
\begin{figure}[htbp]
\begin{center}
\includegraphics[width=15cm]{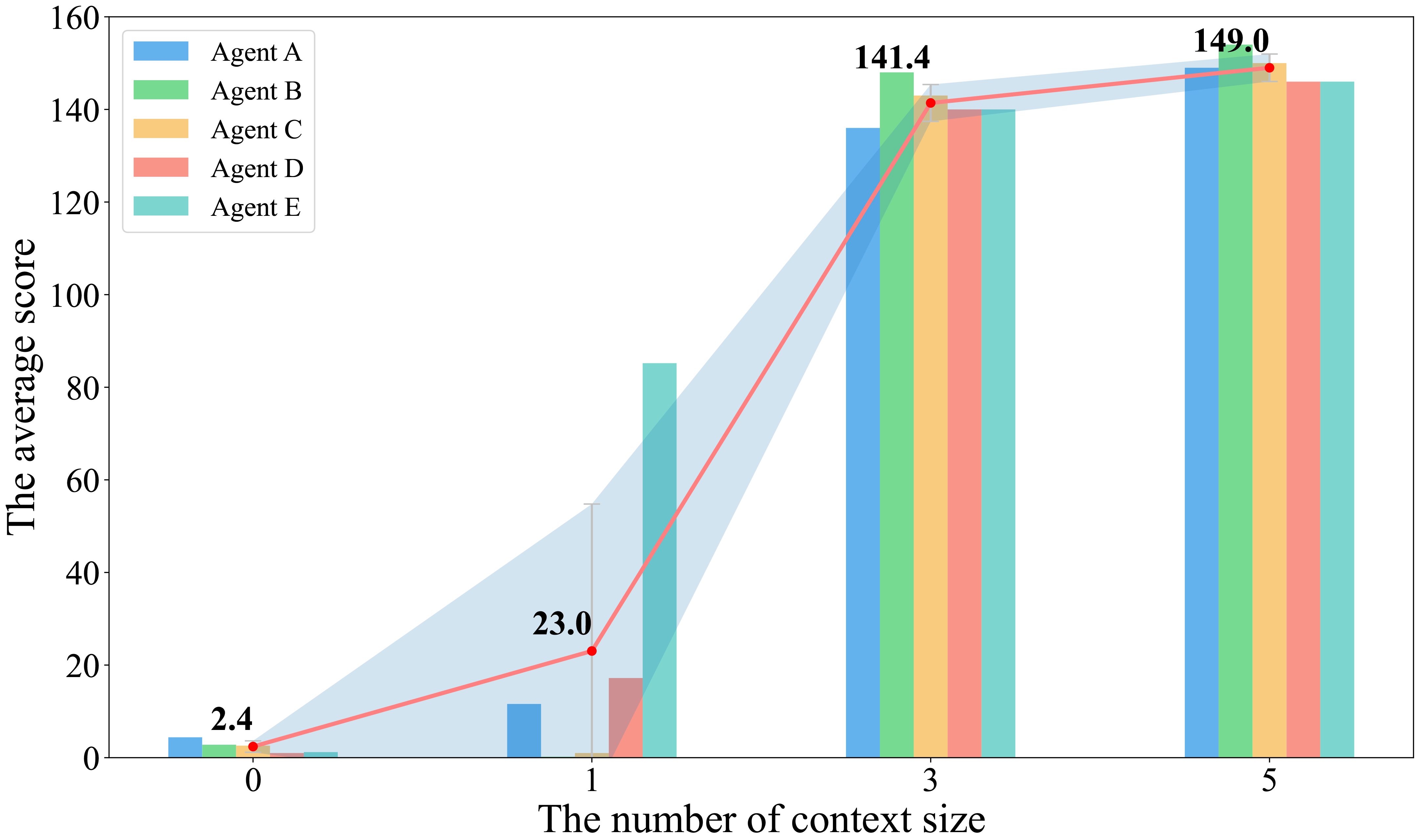}
\end{center}
\caption{This diagram shows the detailed scores for different context sizes (0, 1, 3, 5). A-E denotes different agents with different random seeds. The score improves 
as the number of contexts increases in our method.}
\label{fig_contextsall}
\end{figure}

\subsection{Tables}
\begin{table*}[htb]
\begin{center}
\caption{Mean rewards for all methods.}
\vspace{5pt}
\begin{tabular}{ccccccccc@{}}
\toprule
\textbf{{Methods}}  &  A &  B    & C    & D    & E    & avg      \\ \midrule
\textbf{DNN baseline}  & \textbf{192} & \textbf{202} & \textbf{202} & 167 & \textbf{192} & \textbf{191.0 ($\pm$14.32)}  \\
\textbf{SAN baseline}  & 136 & 154 & 164 & \textbf{173} & 12 & 127.8 ($\pm$66.18) \\
\textbf{MoP-SAN (ours)}    & 149 & 147 & 146 & 150 & 154 & 149.2 ($\pm$3.32)  \\ 
\bottomrule
\end{tabular}
\label{tab:robust_mean_train}
\end{center}
\end{table*}
\begin{table*}[htb]
\caption{Mean value of overall generalization rewards in the generalization phase.}
\begin{center}
\begin{tabular}{ccccccccc@{}}
\toprule
\textbf{{Methods}}  &  A &  B    & C    & D    & E    & avg      \\ \midrule
\textbf{DNN baseline}  & 37.5 & 12 & 53 & 27.5 & 79 & 41.8 ($\pm$25.59)  \\
\textbf{SAN baseline}  & 115.7 & 111.9 & 102 & 85.6 & 143.8 & 112.2 ($\pm$19.11) \\
\textbf{MoP-SAN (ours)}    & \textbf{143} & \textbf{149.3} & \textbf{152} & \textbf{158.8} & \textbf{157} & \textbf{152.0 ($\pm$6.32)}  \\ 
\bottomrule
\end{tabular}
\label{tab:robust_mean_test}
\end{center}
\end{table*}
\begin{table}[htb]
\begin{center}
\caption{The mean score of our method w/o transformer context encoder and our method.}
\vspace{5pt}
\begin{tabular}{ccccccccc@{}}
\toprule
\textbf{{Methods}}  &  A &  B    & C    & D    & E    & avg      \\ \midrule
\textbf{ours w$/$o context encoder}  & 4.4 & 2.8 & 2.6 & 1 & 1.2 & 2.4 ($\pm$1.38) \\
\textbf{ours}    & \textbf{149} & \textbf{154} & \textbf{150} & \textbf{146} & \textbf{146} & \textbf{149 ($\pm$3.32)}  \\
\bottomrule
\end{tabular}
\label{tab:robust_mean_train_contextencoder}
\end{center}
\end{table}
\begin{table}[htb]
\begin{center}
\caption{The mean generalization score of our method w/o DPP module and our method.}
\vspace{5pt}
\begin{tabular}{ccccccccc@{}}
\toprule
\textbf{{Methods}}  &  A &  B    & C    & D    & E    & avg      \\ 
\midrule
\textbf{ours w$/$o DPP module}  & 124 & 131 & 129 & 119.5 & 153.7 & 131.4 ($\pm$13.22) \\
\textbf{ours}    & \textbf{143} & \textbf{149.3} & \textbf{152} & \textbf{158.8} & \textbf{157} & \textbf{152.0 ($\pm$6.32)}  \\
\bottomrule
\end{tabular}
\label{tab:robust_mean_test_dpp}
\end{center}
\end{table}